\documentclass[article,10pt]{IEEEtran}
\usepackage[named]{algo}
\usepackage{graphics}
\usepackage{graphicx}
\usepackage{stfloats}
\usepackage{amssymb}
\usepackage{amsmath}
\usepackage{amsfonts}
\usepackage{subfigure}
\usepackage{float}
\usepackage{pifont}
\usepackage{setspace}
\usepackage{multirow}
\usepackage{url}

\def\QEDclosed{\mbox{\rule[0pt]{1.3ex}{1.3ex}}} 

\def\QED{\QEDclosed} 
\def\proof{\noindent\hspace{0em}{\itshape Proof: }}
\def\endproof{\hspace*{\fill}~\QED\par\endtrivlist\unskip}

\newcommand{\Section}{\section}

\newtheorem{theorem}{Theorem}
\newtheorem{lemma}{Lemma}

\def\urltilda{\kern -.15em\lower .7ex\hbox{\~{}}\kern .04em}
\def\urldot{\kern -.10em.\kern -.10em}
\def\urlhttp{http\kern -.10em\lower -.1ex\hbox{:}\kern -.12em\lower 0ex\hbox{/}\kern -.18em\lower 0ex\hbox{/}}

\newcommand{\ie}{{\it i.e.}}
\DeclareMathOperator{\sign}{sign}
\DeclareMathOperator{\prox}{prox}

\begin{document}

\title{Decreasing Weighted Sorted $\ell_1$ Regularization}
\author{Xiangrong Zeng\ \  and \  M\'{a}rio A. T. Figueiredo  \thanks{Manuscript submitted December 20, 2012. } \\
\thanks{ Both authors are with the Instituto de Telecomunica\c{c}\~oes and the Department of Electrical and
Computer Engineering, Instituto Superior T\'ecnico, University of Lisbon, 1049-001, Lisboa, Portugal.
Email: Xiangrong.Zeng@lx.it.pt, mario.figueiredo@lx.it.pt. }}

\maketitle

\begin{abstract}
We consider a new family of regularizers, termed {\it weighted sorted $\ell_1$ norms} (WSL1), which generalizes the recently introduced {\it octagonal shrinkage and clustering algorithm for regression} (OSCAR) and also contains the $\ell_1$ and $\ell_{\infty}$ norms as particular instances. We focus on a special case of the WSL1, the {\sl decreasing WSL1} (DWSL1), where the elements of the argument vector are sorted in non-increasing order and the weights are also non-increasing. In this paper, after showing that the DWSL1 is indeed a norm, we derive two key tools for its use as a regularizer: the dual norm and the Moreau proximity operator.
\end{abstract}

\Section{Introduction}\label{sec:intro}
In recent years, much research has been devoted not only to sparsity, but also to structured/group sparsity \cite{Bach2012}. The OSCAR \cite{bondell2007simultaneous} is a convex regularizer, which was proposed to
promote variable/feature grouping; unlike other methods, it does not require previous knowledge of the group structure and it is not tied to any particular order of the variables. The OSCAR criterion for linear regression with a quadratic loss function has the form
\begin{equation}\label{OSCAR}
\min_{{\bf x}\in \mathbb{R}^n} \frac{1}{2} \left\| {\bf y} - {\bf A}{\bf x} \right\|_2^2 +
\underbrace{\lambda_1 \left\|{\bf x} \right\|_1 + \lambda_2 \sum_{i<j}
\max \left\{ |x_i |,  |x_j |\right\}}_{\Phi_{\mbox{\tiny OSCAR}} \left( {\bf x}\right)},
\end{equation}
where ${\bf y}\in \mathbb{R}^m$, ${\bf A} \in \mathbb{R}^{m\times n}$, and $\lambda_1, \lambda_2$ are non-negative parameters. The $\ell_1$ norm and the
pairwise $\ell_\infty$ penalty simultaneously encourage sparsity and pair-wise equality (in magnitude), respectively. It is simple to show that he OSCAR regularizer
can be re-written as a weighted and sorted (in non-increasing order)
$\ell_1$ norm,
\begin{equation}\label{OSCAR_reg}
\Phi_{\mbox{\tiny OSCAR}} \left( {\bf x}\right) = \|\grave{\bf e} \odot { \grave{\bf x}} \|_1
 = \sum_{i=1}^{n} \grave{e}_i \; |\grave{x}_i|,
\end{equation}
where $\odot$ is the element-wise multiplication,
$\grave{\bf e} $ is a vector of weights given by
\begin{equation}\label{weightvector}
\grave{e}_i=\lambda_1+ \lambda_2 \left(n-i\right),\;\;\mbox{for}\;i=1,...,n,
\end{equation}
and $\grave{\bf x}$ is a vector obtained from ${\bf x}$ by
sorting its entries in non-increasing order of
magnitude.

The form of \eqref{OSCAR} suggests a possible alternative where
$\max$ is replaced with $\min$, which we call the
{\it small magnitude penalized} (SMAP) regularization (which is non-convex):
\begin{equation}\label{smap}
\min_{\bf x} \frac{1}{2} \left\| {\bf y} - {\bf A}{\bf x} \right\|_2^2 +
\underbrace{\lambda_1 \left\|{\bf x} \right\|_1 + \lambda_2 \sum_{i<j}
{\min} \left\{ |x_i |,  |x_j |\right\}}_{\Phi_{\mbox{\tiny SMAP}} \left( {\bf x}\right)}.
\end{equation}
The SMAP regularizer can also be written as a sorted $\ell_1$  norm,
\begin{equation}\label{SMAP_reg}
\Phi_{\mbox{\tiny SMAP}} \left( {\bf x}\right) = \|\acute{\bf e} \odot { \grave{\bf x}} \|_1
 = \sum_{i=1}^{n} \acute{e}_i \; |\grave{x}_i|,
\end{equation}
where $\acute{\bf e}$ is now a weight vector with elements given by
\begin{equation}\label{weightvector_smap}
\acute{e}_i=\lambda_1+ \lambda_2 \left(i-1\right),\;\;\mbox{for}\;i=1,...,n.
\end{equation}

\begin{figure}
	\centering
		\includegraphics[width=0.85\columnwidth]{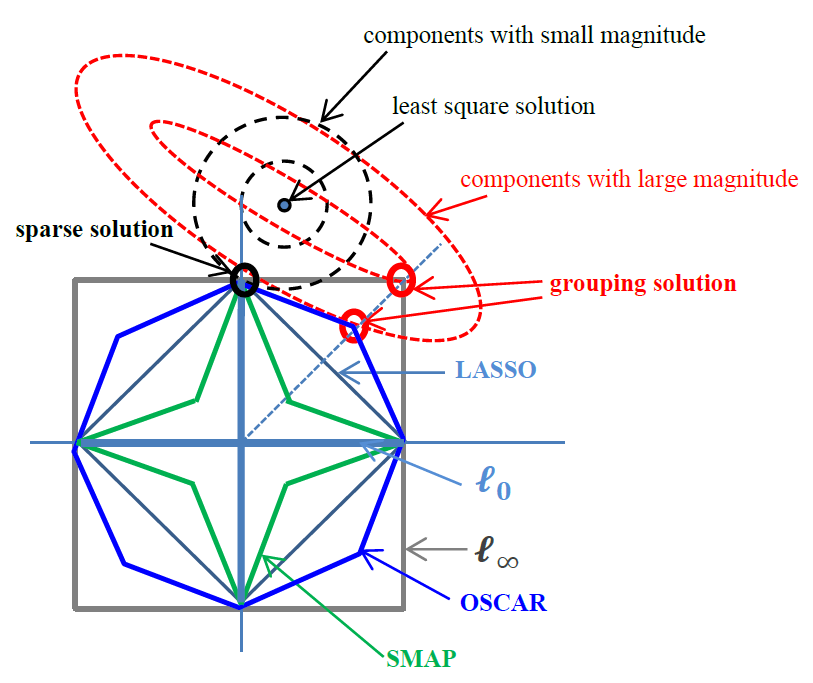}
	\caption{Illustration of $\ell_0$, SMAP, LASSO,
OSCAR and $\ell_{\infty}$}
	\label{fig:figure1}
\end{figure}

It is interesting to observe the relationships among the $\ell_0$ norm, SMAP, LASSO,
OSCAR, and the $\ell_{\infty}$ norm, and the different types of estimates that they
promote. The level curves of several of the regularizers mentioned above
(for the 2D case) are shown in Fig.~\ref{fig:figure1}. The figure illustrates
why these models promote sparsity, grouping, or group sparsity.
A comparison of these regularizers is summarized in Table~\ref{tab:comp_regs}.
Note that the $\ell_0$ norm leads to an NP-hard problem \cite{candes2005decoding},  \cite{natarajan1995sparse};
the SMAP behaves similarly to an $\ell_p$ norm (with $p \in (0, 1)$), and is non-convex; the LASSO is the tightest convex relaxation of the $\ell_0$ norm; the OSCAR is able to
simultaneously promote grouping and sparsity via its equality-inducing (grouping) vertices
and sparsity-inducing vertices, respectively; finally, the $\ell_{\infty}$ norm
only encourages grouping through its equality-inducing vertices.

\begin{table} [h!]
\centering \caption{Comparisons of different regularizers} \label{tab:comp_regs}
\begin{tabular}{|c|c |c|}
\hline
\footnotesize Regularizers   & \footnotesize promoting & \footnotesize convexity \\
\hline
\hline
\footnotesize $\ell_0$        & \footnotesize sparsity & \footnotesize non-convex \\
\footnotesize SMAP          & \footnotesize sparsity     & \footnotesize non-convex \\
\footnotesize LASSO         & \footnotesize sparsity     & \footnotesize convex  \\
 \footnotesize OSCAR        & \footnotesize group sparsity  & \footnotesize convex \\
\footnotesize $\ell_{\infty}$ & \footnotesize grouping     & \footnotesize convex  \\
\hline
\end{tabular}
\end{table}

In this paper, we consider a general formulation that includes, as special cases, the $\ell_1$ and $\ell_{\infty}$ norms, as well as the OSCAR and SMAP regularizers:
the {\it weighted sorted $\ell_1$} (WSL1) norm, given by
\begin{equation}
\Omega_{{\bf t}} \left({\bf x}\right)  = \left\|{\bf t} \odot \grave{\bf x}\right\|_1 ,
\end{equation}
where ${\bf t}$ is a non-negative weight vector and $\grave{\bf x}$ is
as defined above. We consider first that the entries of ${\bf t}$ form a non-increasing sequence,
$t_1 \geq \cdots \geq t_n \geq 0$, which includes the following notable cases:
\begin{itemize}
	\item if ${\bf t} = \grave{\bf e}$ (as given by \eqref{weightvector}), then $\Omega_{{\bf t}} ({\bf x})
	= \Phi_{\mbox{\tiny OSCAR}} ( {\bf x} )$;
	\item if ${\bf t} = \grave{\bf e}$ with $\lambda_2 = 0$, then $\Omega_{{\bf t}} ({\bf x}) = \lambda_1 \|{\bf {x}} \|_1$;
    \item if $t_1 > 0, $ and $t_i = 0$, for $i \geq 2,$ then $\Omega_{{\bf t}} ({\bf x}) = t_1 \|{\bf {x}} \|_{\infty}$.
\end{itemize}

We consider now cases where the entries of ${\bf t}$ form a non-decreasing sequence, $0 \leq t_1 \leq \cdots \leq t_n$, which includes the following notable cases:
\begin{itemize}
	\item if ${\bf t} = \acute{\bf e}$, then $\Omega_{{\bf t}} ({\bf x})
	= \Phi_{\mbox{\tiny SMAP}} ( {\bf x})$;
	\item if ${\bf t} = \acute{\bf e}$ with $\lambda_2 = 0$, then $\Omega_{{\bf t}} ({\bf x}) = \lambda_1 \|{\bf {x}} \|_1$.
\end{itemize}

As shown below, $\Omega_{{\bf t}}$ is indeed a norm, as long as the components of ${\bf t}$ form a non-increasing (and non-zero) sequence, and is able to promote group sparsity if the entries of ${\bf t}$ form a decreasing sequence. In this paper, we focus on this case, termed {\it decreasing WSL1} (DWSL1); we analyze its nature, obtain its dual norm and its Moreau proximity operator, which is fundamental building block for its use as a regularizer.


While we were preparing this manuscript, we became aware of the proposal of an {\it ordered $\ell_1$ norm} \cite{bogdan2013statistical} for
statistical estimation and testing, which is formally equivalent to the DWSL1, and also a special case of the WSL1, although it was motivated from different considerations. Moreover, one of the central results in this paper (Theorem \ref{the:theorem2}) is an extension of our previous work \cite{zeng2013solving}, which predates \cite{bogdan2013statistical}.

\Section{Formulation, Dual Norm and Optimality Conditions}

\subsection{Proposed Formulation}

The DWSL1-regularized least square problem is
\begin{equation} \label{dwsl1}
\min_{{\bf x} \in \mathbb{R}^n}
\tfrac{1}{2} \left\| {\bf y} - {\bf A}{\bf x} \right\|_2^2 +  \Omega_{\grave{\bf w}} \left({\bf x}\right),
\end{equation}
where $\Omega_{\grave{\bf w}}$ is the DWSL1 norm, defined as
\begin{equation}
\Omega_{\grave{\bf w}}:\mathbb{R}^n \rightarrow \mathbb{R}, \;\;\; \Omega_{\grave{\bf w}}({\bf x}) = \left\|\grave{\bf w} \odot \grave{\bf x}\right\|_1
\end{equation}
(with $\odot$ denoting the component-wise product), $\grave{\bf x}$ is a vector obtained from ${\bf x}$ by
sorting its entries in non-increasing order of magnitude (with ties broken by an arbitrary fixed rule) and ${\bf w}$ is a vector of weights such that its components form a non-increasing sequence:
\[
\grave{w}_1 \geq \grave{w}_2 \geq \cdots \geq \grave{w}_n.
\]

Let ${\bf P}({\bf x})$ be the permutation matrix that sorts ${\bf x}$ into $\grave{\bf x}$, {\it i.e.},
\begin{equation}
\grave{\bf x} = {\bf P}({\bf x}) \, {\bf x}, \label{x_sort}
\end{equation}
which, being a permutation matrix, satisfies the equality $\bigl({\bf P}({\bf x})\bigr)^{-1} = \bigl({\bf P}({\bf x})\bigr)^{T}$. We can then write
\begin{equation}
\Omega_{\grave{\bf w}}({\bf x}) = \|\grave{\bf w}\odot \bigl({\bf P}({\bf x})\, {\bf x}\bigr)\|_1 = \|\bigl(({\bf P}({\bf x}))^T\,\grave{\bf w}\bigr) \odot  {\bf x}\|_1.
\end{equation}

The convexity of $\Omega_{\grave{\bf w}}$ and the fact that it is a norm are the subject of the following two lemmas. We should point out that similar results were recently proved (using different arguments) in \cite{bogdan2013statistical}.

\vspace{0.1cm}
\begin{lemma} \label{lemma2} $\Omega_{\grave{\bf w}}:\mathbb{R}^n \rightarrow \mathbb{R}$ is a convex function.
\end{lemma}
\vspace{0.1cm}
\proof
Let ${\bf u}, {\bf v} \in \mathbb{R}^n$, $\theta \in [0,1]$,
${\bf x} = \theta{\bf u} + \left(1- \theta\right){\bf v}$,  then
$\Omega_{\grave{\bf w}} \left({\bf u}\right) = \left\|\grave{\bf w} \odot \grave{\bf u}\right\|_1,
\Omega_{\grave{\bf w}} \left({\bf v}\right) = \left\|\grave{\bf w} \odot \grave{\bf v}\right\|_1,
\Omega_{\grave{\bf w}} \left( {\bf x}\right)
= \left\|\grave{\bf w} \odot \grave{\bf x}\right\|_1$,
where $\grave{\bf u} = {\bf P}\left({\bf u}\right){\bf u}$,
$\grave{\bf v} = {\bf P}\left({\bf v}\right){\bf v}$,
and
$\grave{\bf x} = {\bf P}\left({\bf x}\right){\bf x}$.
Thus,
\[
\begin{split}
\Omega_{\grave{\bf w}} \left( {\bf x}\right) & = \|\bigl({\bf P}({\bf x}) {\bf x}\bigr) \odot \grave{\bf w} \|_1\\
& = \|\bigl({\bf P}({\bf x}) (\theta{\bf u} + (1-\theta){\bf v})\bigr)\odot \grave{\bf w} \|_1 \\
& \stackrel{(a)}{\leq}
\theta \; \|\bigl({\bf P}({\bf x}) {\bf u}\bigr) \odot \grave{\bf w} \|_1
+ (1-\theta)\; \|\bigl( {\bf P}({\bf x}) {\bf v}\bigr)\odot \grave{\bf w} \|_1\\
& \stackrel{(b)}{\leq}
\theta \; \|{\bf P}({\bf u}) {\bf u} \odot \grave{\bf w} \|_1
+ (1-\theta)\; \|\bigl({\bf P}({\bf v}) {\bf v}\bigr) \odot \grave{\bf w} \|_1\\
& = \theta \; \Omega_{\grave{\bf w}} \left({\bf u}\right) + (1-\theta)\; \Omega_{\grave{\bf w}} \left({\bf v}\right),
\end{split}
\]
where $(a)$ is simply the triangle inequality together with the positive homogeneity of norms ({\it i.e.}, $\|\alpha {\bf x}\|_1 = |\alpha|\, \|{\bf x}\|_1$, for any $\alpha\in\mathbb{R}$) and $(b)$ results from the following fact: if the components of a vector ${\bf b}$ form a non-increasing non-negative sequence, then $\left\|{\bf P}({\bf c}){\bf a} \odot {\bf b}\right\|_1 \leq \|{\bf P}({\bf a}) {\bf a} \odot {\bf b}\|_1$, for any ${\bf a}$, ${\bf c}$. \endproof
\vspace{0.1cm}

\begin{lemma} If $\grave{w}_1 > 0$, then $\Omega_{\grave{\bf w}}:\mathbb{R}^n \rightarrow \mathbb{R}$ is a norm.
\end{lemma} \label{lemma2b}

\proof The positive homogeneity of $\Omega_{\grave{\bf w}}$  is obvious (and was in fact already used in the proof of Lemma \ref{lemma2}). The triangle inequality results trivially from the convexity shown in Lemma \ref{lemma2}, by taking $\theta = \tfrac{1}{2}$, combined with the positive homogeneity. Finally, we need to prove that $\Omega_{\grave{\bf w}}({\bf x}) = 0 \Leftrightarrow {\bf x} = 0$; this is true if $\grave{w}_1 > 0$, since it is clear that $\Omega_{\grave{\bf w}}({\bf x}) \geq \grave{w}_1 \|{\bf x}\|_{\infty}$.
\endproof

\vspace{0.15cm}

Although not needed for these proofs, it is also clear that $\grave{w}_1 \|{\bf x}\|_1 \geq \Omega_{\grave{\bf w}}({\bf x})$.

\subsection{Dual Norm}
We will now obtain the dual norm of $\Omega_{\grave{\bf w}}$, which, by definition, is given by
\begin{equation} \label{dual_norm}
\Omega^*_{\grave{\bf w}} ({\bf x}) =
\max_{{\bf u} \in {\cal S}} \left\langle {\bf u}, {\bf x}\right\rangle ,
 \end{equation}
 where ${\cal S} = \{{\bf u}:  \Omega_{\grave{\bf w}} ({\bf u}) \leq 1\}$.
Notice that ${\cal S}$ is a closed bounded convex set, since it corresponds to the unit ball of a norm, and a polytope, since it is defined by a finite number of linear inequalities.  According to a well-known fundamental result in linear programming, the maximum (and minimum) of a linear function over a bounded closed convex polygonal region is attained at least at one of the region's vertices. We may thus focus only on the vertices of ${\cal S}$ to compute $\Omega^*_{\grave{\bf w}}$.
\begin{figure}
	\centering
		\includegraphics[width=0.70\columnwidth]{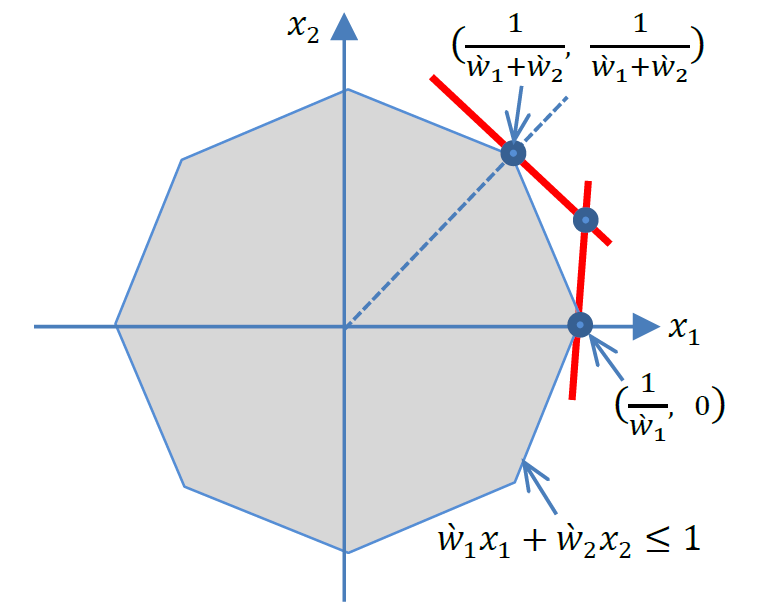}
	\caption{Illustration of computing $\Omega^*_{\grave{\bf w}} \left({\bf x}\right)$ in
the 2-dimensional case.}
	\label{fig:figure2}
\end{figure}

Since it is clear that $\Omega_{\grave{\bf w}} ({\bf u})$ depends only on the magnitudes of the components of ${\bf u}$ we may limit our attention to the first orthant. Moreover, since $\Omega_{\grave{\bf w}} ({\bf u})$ is invariant to permutations of the components of ${\bf u}$ (because of the underlying sorting operation), we may consider, without loss of generality, some vertex ${\bf c} = [c_1,c_2,...,c_n]$, satisfying
\begin{equation}
c_1 \geq c_2 \geq \cdots \geq c_n \geq 0.\label{eq:sorted_entries}
\end{equation}

The vertices of a convex polyhedron are those points in the polyhedron that can not be obtained as convex combinations of other points; that is, such that the set is the convex hull of that set of points. In the case of ${\cal S}$, the subset of vertices in the first orthant, with components satisfying \eqref{eq:sorted_entries}, is
\begin{equation}
\tilde{\cal C} = \{{\bf c}_1, {\bf c}_2, \cdots, {\bf c}_n\},
\end{equation}
with
\begin{equation}
\begin{split}
{\bf c}_1 &= [\tau_1, 0, 0, \cdots, 0 ]^T \\
{\bf c}_2 &= [\tau_2, \tau_2, 0, \cdots, 0 ]^T \\
\vdots\\
{\bf c}_{n} &= [\tau_{n}, \tau_{n}, \cdots,\tau_{n}, \tau_{n}]^T \\
\end{split}
\end{equation}
where
\begin{equation}
\tau_i = \left( \sum_{j=1}^i \grave{w}_j\right)^{-1}\!\!, \;\; i \in \{1, \cdots, n \}.
\end{equation}
It is clear that these points belong to ${\cal S}$ and that none of them can be obtained as a convex combination of the others. Figure \ref{fig:figure2} depicts the 2-dimensional case. The complete set of vertices ${\cal C}$ (in the general case, see the next paragraph for exceptions) is obtained by considering all possible permutations of the entries of each element of $\tilde{\cal C}$ and all possible sign configurations.

Notice that for certain particular choices of $\grave{\bf w}$, some of the points mentioned in the previous paragraph are no longer vertices. For example, if $\grave{w}_1 = ... = \grave{w}_2 = w$, we have $\tau_i = (i w)^{-1}$; in this case, only the points that have exactly one non-zero component are vertices, since in this case we recover the $\ell_1$ norm.

Given some vector ${\bf x}\in \mathbb{R}^n$, let ${\bf x}_{(k)}$ denote the vector obtained by keeping the $k$ largest (in magnitude) elements of its sorted version $\grave{\bf x}$ and setting the others to zero.
Also, let $x_{(k)}$ be the $k$-largest element of ${\bf x}$ in absolute value.
Naturally, $\left\| {\bf x}_{(1)}\right\|_1 = |x_{(1)}| = \left\|{\bf x}\right\|_{\infty} $
and $\left\| {\bf x}_{(n)}\right\|_1 = \left\|{\bf x}\right\|_1$.
Therefore, we have
\begin{equation} \label{dual_norm}
\begin{split}
\Omega^*_{\grave{\bf w}} \left({\bf x}\right) &=
\max_{{\bf u} \in {\cal S}} \left\langle {\bf u}, {\bf x}\right\rangle\\
& = \max_{{\bf c} \in {\cal C}} \left\langle {\bf c},\grave{\bf x}\right\rangle \\
&= \max \{ \langle {\bf c}_k,|{\bf x}_{(k)}|\rangle, k =1, \cdots, n \},
\end{split}
 \end{equation}
where $|{\bf v}|$ denotes the vector obtained by taking the absolute values of the components of ${\bf v}$. The derivation in \eqref{dual_norm} proves the following theorem.

\vspace{0.15cm}
\begin{theorem} The dual norm of $\Omega_{\grave{\bf w}}$ is given by
\begin{equation} \label{dual_norm_final}
\Omega_{\grave{\bf w}}^* ({\bf x}) = \max \Bigl\{\tau_k \bigl\|{\bf x}_{(k)}\bigr\|_1,\;\; k = 1, \cdots, n\Bigr\}.
 \end{equation}
\end{theorem}
\vspace{0.15cm}

As a simple corollary of this theorem, notice that, in the case where $\grave{w}_1 = ... = \grave{w}_2 = w$ (that is, if $\Omega_{\grave{\bf w}} ({\bf x}) = w\|{\bf x}\|_1$), it results from \eqref{dual_norm_final} that
\begin{equation} \label{dual_norm_l1_linf}
\Omega_{\grave{\bf w}}^* ({\bf x})  = \max \biggl\{ \frac{\bigl\|{\bf x}_{(k)}\bigr\|_1}{k\, w} ,\;\; k = 1, \cdots, n\biggr\} = \frac{1}{w} \, \| {\bf x}\|_{\infty}
\end{equation}
because
\[
|x_{(1)}| \geq \tfrac{1}{2}(|x_{(1)}|+|x_{(2)}| ) \geq ... \geq
\tfrac{1}{n}(|x_{(1)}|+ ... + |x_{(n)}| ),
\]
recovering the well-known fact that the dual of the $\ell_1$ norm is the $\ell_{\infty}$ norm.

\section{Proximity Operator}
A fundamental building block for using the DWSL1 norm as a regularizer, since it opens the door to the use of proximal gradient algorithm for solving \eqref{dwsl1}, is its Moreau proximity operator, defined as
\begin{equation}\label{POofDWSL1}
\prox_{\Omega_{\grave{\bf w}}} \left( {\bf v}\right) =
\arg\min_{{\bf x}\in \mathbb{R}^n} \left( \Omega_{\grave{\bf w}} \left({\bf x}\right)  + \frac{1}{2} \left\|{\bf x}-{\bf v}\right\|^2\right).
\end{equation}

In this section, we now show how to compute \eqref{POofDWSL1}, by exploiting the fact that DWSL1 is a generalization of the OSCAR regularizer, extending previous results for OSCAR presented in \cite{zeng2013solving} and \cite{zhong2012efficient}.

The following lemma is a simple generalization of Lemma 1 from \cite{zeng2013solving}:
\vspace{0.1cm}
\begin{lemma}
\label{lem:lemma1}
Consider ${\bf v} \in \mathbb{R}^{n }$, ${\bf P}({\bf v})$, and
${\bf \grave{v}} = {\bf P(v)\, v}$ as defined above.
Let \begin{equation}\label{getmustar}
{\bf u}^* = \arg\min_{{\bf u}\in \mathbb{R}^n} \left( \|{\bf \boldsymbol{\pi}} \odot {\bf u} \|_1 +
\frac{1}{2} \left\|{\bf u}-{\bf \grave{v}} \right\|^2 \right)
\end{equation}
where ${\bf \boldsymbol{\pi}}\in \mathbb{R}_+^n$ is a positive weight vector. If
\begin{equation}\label{keycondition}
|\grave{v}_i| - {\pi}_i \geq  |\grave{v}_{i+1}| -
{\pi}_{i+1}, \;\; \mbox{for}\; i=1,\dots,n-1,
\end{equation}
then ${\bf u}^*$  satisfies $|u^*_i|\geq |u^*_{i+1}|$, for  $i=1,2,...,n-1$.
Moreover, if \eqref{keycondition} is satisfied with ${\bf \boldsymbol{\pi}} = \grave{\bf w} = {\bf P}({\bf v}){\bf w}$,  then $\prox_{\Omega_{\grave{\bf w}}} ( {\bf v})= \bigl({\bf P}({\bf v})\bigr)^T {\bf u}^* $.
\end{lemma}
\vspace{0.1cm}

We next describe the proximity operator of $\Omega_{\grave{\bf w}}$, based on the
results from \cite{zhong2012efficient} and the analysis in \cite{zeng2013solving}, and
leveraging Lemma \ref{lem:lemma1}. Begin by noting that, since $\Omega_{\grave{\bf w}}({\bf x})
= \Omega_{\grave{\bf w}}(|{\bf x}|)$ and, for any ${\bf x},{\bf v} \in \mathbb{R}^n$ and ${\bf u} \in \{-1,\, 0,\, 1\}^n$,
\[
\|\mbox{sign}({\bf v})\odot |{\bf x}| - {\bf v}\|_2^2  \leq \| {\bf u} \odot |{\bf x}| - {\bf v}\|_2^2,
\]
we have that
\begin{equation} \label{signfunction}
\mbox{sign}\left(\prox_{\Omega_{\grave{\bf w}}} \left( {\bf v}\right) \right) = \mbox{sign}({\bf v}).
\end{equation}
Consequently, we have
\begin{equation}\label{xtarfromx}
\prox_{\Omega_{\grave{\bf w}}}({\bf v}) = \mbox{sign}\left({\bf v}\right) \odot \prox_{\Omega_{\grave{\bf w}}}(|{\bf v}|),
\end{equation}
showing that there is no loss of generality in assuming ${\bf v} \geq 0$, \ie, the fundamental step in computing $\prox_{\Omega_{\grave{\bf w}}}({\bf v})$ consists in obtaining
\begin{equation}\label{a_with_v}
\begin{split}
{\bf a} &= \arg\min_{{\bf x}\in \mathbb{R}_+^n} \left( \Omega_{\grave{\bf w}}\left({\bf x}\right) +
\frac{1}{2} \left\|{\bf x}- |{\bf v}|\right\|^2 \right).
 \end{split}
\end{equation}
Furthermore, both terms in the objective function in
\eqref{a_with_v} are invariant under a common permutation of the components of
the vectors involved; thus,
denoting
\begin{equation}\label{b_with_vsort}
{\bf b} = \arg\min_{{\bf x}\in \mathbb{R}_+^n } \left( \Omega_{\grave{\bf w}}\left({\bf x}\right)
+ \frac{1}{2} \left\|{\bf x}- |{ \grave{\bf v}}|\right\|^2 \right),
\end{equation}
where (as defined above) $ \grave{\bf v} = {\bf P}({\bf v})\; {\bf v}$,  allows writing
\begin{equation} \label{a_and_b}
{\bf a} = \bigl({\bf P}({\bf v})\bigr)^T{\bf b},
\end{equation}
showing that there is no loss of generality in assuming that
the elements of ${\bf v}$ are sorted in decreasing magnitude.
As shown in the following theorem,
${\bf b}$  has several important properties.

\vspace{0.2cm}
\begin{theorem}
\label{the:theorem1}
 Letting ${\bf b}$ be as defined in \eqref{b_with_vsort}, we have that
\begin{itemize}
	\item [(i)] For  $i=1,2,...,n-1$, $b_i\geq b_{i+1}$;
	moreover,
  \[(|\grave{v}_p| = |\grave{v}_{q}|) \Rightarrow ( b_p= b_q).
  \]
	\item [(ii)] The property of ${\bf b}$ stated in (i) allows
   writing it as
   \[{\bf b} =
[{ b}_1 ... { b}_n]^T
= \left[{ b}_{s_1}...
{ b}_{t_1}...{ b}_{s_j}...
{ b}_{t_j}...
{ b}_{s_l}...{ b}_{t_l} \right]^T,
\]
where $b_{s_j} = \cdots = b_{t_j}$ is the $j$-th group of consecutive
equal elements of ${\bf b}$ (of course, ${s_1} = 1$ and $t_l = n$)  and there are $l \in \{1,.., n\}$ such
groups. For the $j$-th group, the common optimal value is \begin{equation}\label{average}
{ b}_{s_j}=\cdots={ b}_{t_j} =
\max \left\{{ \bar{v}}_j - \bar{w}_j, 0 \right\}
\end{equation}
where
\begin{equation} \label{zbarandwbar1}
 { \bar{v}}_j =  \frac{1}{\vartheta_j}\sum_{i = s_j}^{t_j}|\grave{v}_{i}|,
\end{equation}
is the $j$-th group average (with ${\vartheta_j} = t_j - s_j + 1$ denoting its number of components) and
\begin{equation}  \bar{w}_j  =
 \frac{1}{\vartheta_j} \sum_{i = s_j}^{t_j} \grave{w}_{i}.\label{zbarandwbar2b}
\end{equation}
	\item [(iii)] For the $j$-th group, if ${ \bar{v}}_j - \bar{w}_j \geq 0 $, then
	there is no integer $r \in \left\{s_j,\cdots,t_j-1\right\}$
	such that \[
    \sum_{i=s_j}^r |\grave{v}_i|- \sum_{i=s_j}^r \grave{w}_i > 	
    \sum_{i=c+1}^{t_j} |\grave{v}_i|- \sum_{i=c+1}^{t_j} \grave{w}_i.
\] Such a group is called {\rm coherent}, since it cannot be split
into two groups with different values that respect (i).
\end{itemize}
 \end{theorem}
\vspace{0.1cm}

Theorem \ref{the:theorem1} is a generalization of Theorem 1 in \cite{zeng2013solving}, and Theorem 1 and Propositions 2, 3, and 4 in \cite{zhong2012efficient}, where an algorithm was also proposed to obtain the optimal ${\bf b}$. That algorithm equivalently divides the indices of $|{\bf \grave{v}}|$ into
groups and performs averaging within each group (according to \eqref{zbarandwbar1}),
obtaining a vector that we denote as ${\bf \bar{v}}$; this operation of grouping
and averaging is denoted as
\begin{equation}\label{groupandaverage}
\bigl(\bar{\bar{{\bf v}}},\bar{\bar{{\bf w}}}\bigr) = \mbox{GroupAndAverage}\left(|{\grave{\bf v}}|,{\bf w}\right),
\end{equation}
where
\begin{equation}\label{newzsort}
\bar{\bar{{\bf v}}} = [
\underbrace{{ \bar{v}}_1,\ldots,{ \bar{v}}_1}_{\vartheta_1 \; \mbox{\footnotesize components}}
\ldots \underbrace{{ \bar{v}}_j\ldots{ \bar{v}}_j}_{\vartheta_j  \; \mbox{\footnotesize components}}
\ldots \underbrace{{ \bar{v}}_l\ldots{ \bar{v}}_l}_{\vartheta_l \; \mbox{\footnotesize components}} ]^T,
\end{equation}
and
\begin{equation}\label{newwsort}
\bar{\bar{{\bf w}}} = [
\underbrace{{ \bar{w}}_1,\ldots,{ \bar{w}}_1}_{\vartheta_1 \; \mbox{\footnotesize components}}
\ldots \underbrace{{ \bar{w}}_j\ldots{ \bar{w}}_j}_{\vartheta_j  \; \mbox{\footnotesize components}}
\ldots \underbrace{{ \bar{w}}_l\ldots{ \bar{w}}_l}_{\vartheta_l \; \mbox{\footnotesize components}} ]^T,
\end{equation}
with the $\bar{v}_j$ as given in \eqref{zbarandwbar1} and
the $\bar{w}_j$ as given in \eqref{zbarandwbar2b}.
Finally, ${\bf b}$ is obtained as
\begin{equation}\label{equivalentform}
{\bf b} = \max(\bar{\bar{{\bf v}}}-\bar{\bar{\bf  w}}, 0 ).
\end{equation}

The following lemma, which is a simple corollary of Theorem \ref{the:theorem1},
indicates that condition \eqref{keycondition} is satisfied with $\boldsymbol{\pi} = \bar{\bar{\bf  w}}$.

\vspace{0.1cm}
\begin{lemma}
\label{lem:lemma2}
Vectors $\bar{\bar{{\bf v}}}$ and $\bar{\bar{\bf  w}}$ satisfy
\begin{equation}
\bar{\bar{v}}_i - \bar{\bar{w}}_i \geq  \bar{\bar{v}}_{i+1} -
\bar{\bar{w}}_{i+1}, \;\; \mbox{for}\; i=1,\dots,n-1.
\end{equation}
 \end{lemma}
\vspace{0.1cm}

We are now ready to give the following theorem for $\mbox{prox}_{\Omega_{\grave{\bf w}}} ( {\bf v})$,
(which generalizes Theorem 2 from \cite{zeng2013solving}):

\vspace{0.1cm}
\begin{theorem}
\label{the:theorem2}
Consider ${\bf v} \in \mathbb{R}^{n }$ and a permutation matrix ${\bf P}({\bf v})$
such that the elements of ${\grave{\bf v}} = {\bf P}({\bf v})\, {\bf v}$ satisfy
$|\grave{v}_i| \geq |\grave{v}_{i+1}|$, for $i=1,2,...,n-1$.
Let ${\bf a}$ be given by \eqref{a_and_b}, where ${\bf b}$ is given by
\eqref{equivalentform}; then
$\prox_{\Omega_{\grave{\bf w}}} ( {\bf v}) = {\bf a}\odot \sign({\bf v})$.
\end{theorem}
\vspace{0.2cm}

The fact that the proximity operator of DWSL1 can be efficiently obtained makes
solving \eqref{dwsl1} possible via state-of-the-art proximal gradient algorithms, such as FISTA \cite{beck2009fast}, TwIST \cite{bioucas2007new}, or SpaRSA \cite{wright2009sparse}, or by using the alternating direction method of multipliers (ADMM) \cite{boyd2011distributed} or the split-Bregamn method (SBM) \cite{goldstein2009split}. The convergences of the aforementioned algorithms are guaranteed by their own convergence results,
since the proximity operator stated above is exact.

\Section{Conclusions}
\label{sec:conclusions}
In this short paper, we have presented some results concerning the so-called {\it decreasing weighted sorted $\ell_1$ norm} (DWSL1), which is an extension of the OSCAR ({\it octagonal shrinkage and clustering algorithm for regression} \cite{bondell2007simultaneous}) that are fundamental building blocks for its use and analysis as a regularizer. Namely, after showing that the DWSL1 is in fact a norm, we have derived its dual norm and presented its Moreau proximity operator.

\bibliographystyle{IEEEtran}
\bibliography{bibfile_spl}

\end{document}